%% file: paper.tex
\documentclass[]{fairmeta}

\title{Inpainting-Guided Policy Optimization for Diffusion Large Language Models}

\author[1,2,*, \dagger]{Siyan Zhao}
\author[1]{Mengchen Liu}
\author[1]{Jing Huang}
\author[3]{Miao Liu}
\author[1, 4]{Chenyu Wang}
\author[1]{Bo Liu}
\author[1]{Yuandong Tian}
\author[1]{Guan Pang}
\author[1]{Sean Bell}
\author[2]{Aditya Grover}
\author[1, \dagger]{Feiyu Chen}

\affiliation[1]{Meta Superintelligence Labs}
\affiliation[2]{UCLA}
\affiliation[3]{Tsinghua University, College of AI}
\affiliation[4]{MIT}

\contribution[*]{Work done at Meta}
\contribution[\dagger]{Core Contribution}

\input{sections/abstract}

\date{\today}
\correspondence{Siyan Zhao at \email{siyanz@cs.ucla.edu}, Feiyu Chen at \email{feiyuc@meta.com}}

\begin{document}

\maketitle

\input{sections/intro}

\input{sections/background}

\input{sections/method}
\input{sections/experiments}
\input{sections/related_works}
\input{sections/conclusion}

\clearpage
\newpage
\bibliographystyle{assets/plainnat}
\bibliography{iclr2026/citation}

\clearpage
\newpage
\input{sections/appendix}

\end{document}

%% file: sections/abstract.tex
\abstract{
Masked diffusion large language models (dLLMs) are emerging as promising alternatives to autoregressive LLMs, offering competitive performance while supporting unique generation capabilities such as \emph{inpainting}. We explore how \emph{inpainting} can inform RL algorithm design for dLLMs. Aligning LLMs with reinforcement learning faces an exploration challenge: sparse reward signals and sample waste when models fail to discover correct solutions. While this inefficiency affects LLMs broadly, dLLMs offer a distinctive opportunity—their inpainting ability can guide exploration. We introduce IGPO (Inpainting Guided Policy Optimization), an RL framework that strategically inserts partial ground-truth reasoning traces during online sampling. Unlike providing full solutions, inpainting steers exploration toward promising trajectory spaces while preserving self-generated reasoning, bridging supervised fine-tuning and reinforcement learning.
We apply IGPO to group-based optimization methods such as GRPO, where exploration failures cause zero advantages and gradients. IGPO restores meaningful gradients while improving sample efficiency. We also propose supervised fine-tuning on synthetically rewritten concise traces that better align with dLLM generation patterns. With additional techniques including entropy-based filtering, our training recipe yields substantial gains across three mathematical benchmarks—GSM8K, Math500, and AMC—achieving new state-of-the-art results for full-attention masked dLLMs.
}

%% file: sections/intro.tex
\begin{figure}[H]
    \centering
    \includegraphics[width=\columnwidth]{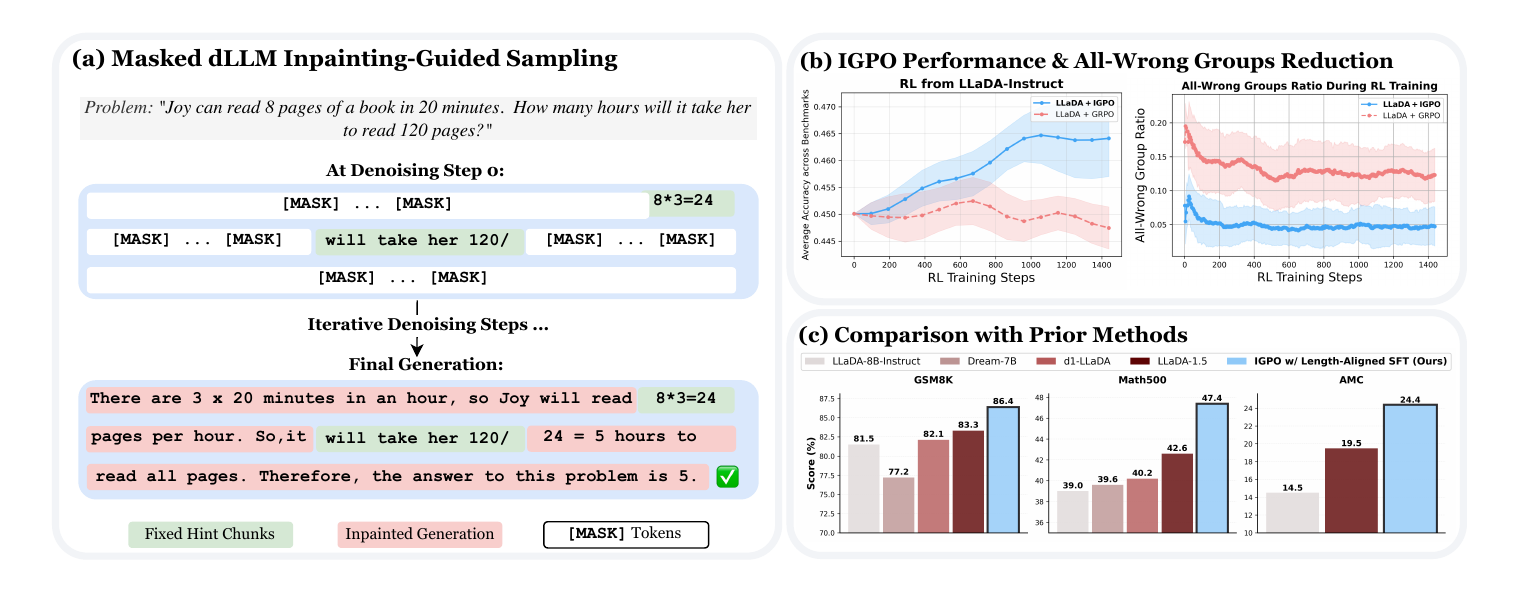}
    \caption{(a) Unlike autoregressive LLMs, diffusion LLMs can be conditioned on future reasoning hints during generation through \emph{inpainting} via bidirectional attention, enabling guided exploration toward correct solutions. (b) Applying inpainting-guided exploration in policy optimization outperforms standard Group-relative Policy Optimization (GRPO) sampling and reduces all-wrong groups occurrences. (c) Our full training recipe combining \emph{Length-Aligned} supervised fine-tuning on concise reasoning traces with IGPO achieves SoTA performance among full-attention masked dLLMs across three mathematical benchmarks.}
    \label{fig:teaser}
\end{figure}

\section{Introduction}

Recent research has shown that masked diffusion large language models (dLLMs)~\citep{austin2021structured,loudiscrete, shi2024simplified} such as LLaDA~\citep{nie2025largelanguagediffusionmodels} and Dream~\citep{dream2025} can achieve performance competitive with autoregressive models of similar size. Moreover, their capabilities and performance can be further enhanced via RL post-training~\citep{zhao2025d1} and ability to flexibly include multimodal data~\citep{li2025lavida, yang2025mmada,you2025llada}.
Unlike autoregressive models, which decode in a left-to-right manner, dLLMs iteratively unmask tokens in parallel. This formulation brings potential for faster inference as shown in closed models like Mercury~\citep{inception2025mercury} and Gemini Diffusion~\citep{deepmind2025gemini}, along with a flexible inductive bias for operations such as \emph{inpainting}, the ability to fill in missing content within existing text.

In this work, we explore how \emph{inpainting} can be leveraged to inform post-training algorithms for dLLMs. Recent work on post-training alignment of dLLMs has adopted training approaches similar to autoregressive LLMs, applying Reinforcement Learning with Verifiable Reward (RLVR) methods and demonstrating promising results across reasoning tasks~\citep{zhao2025d1, yang2025mmada, gong2025diffucoder}. However, a fundamental exploration challenge persists: for challenging tasks, policies struggle to discover correct solutions and binary rewards provide minimal learning signal when most generated solutions are incorrect. This leads to substantial sample waste and poor training efficiency, exacerbating the computational costs of online RL.

The bidirectional generative structure of diffusion models provides a unique mechanism to address this exploration challenge. Since dLLMs are trained through stochastic masking patterns, they possess inherent capability for accepting externally provided partial hints through inpainting operations. We leverage this ability to introduce \inpaint{} (Inpainting Guided Policy Optimization), a novel RL framework that strategically guides exploration for dLLMs by injecting reasoning hints when answering difficult problems. Specifically, when the policy is unlikely to generate correct solutions, partial reasoning traces are injected into the generation region, and the dLLM is tasked with completing the remaining reasoning sequence and output final answer. The final answers are verified against ground truth, and only successful completions are used for downstream policy optimization to mitigate the non-positive learning signal problem.

We demonstrate \inpaint{}'s effectiveness by applying it to group-based policy optimization methods such as Group-relative Policy Optimization (GRPO)~\citep{shao2024deepseekmath}. These methods are particularly vulnerable to exploration failures: when all sampled responses within a group produce uniformly incorrect (or correct) rewards, advantage estimation relies on group-relative reward normalization, causing the advantage signal to collapse to zero and resulting in zero gradients. This phenomenon occurs with alarming frequency in challenging reasoning domains, making group-based RL methods an ideal testbed for our approach. By reducing the prevalence of all-wrong groups, \inpaint{} restores non-degenerate gradient signals, accelerates convergence and enables more effective RL. 

More broadly, \inpaint{} can be viewed as a form of \emph{guided exploration} that interpolates between supervised and reinforcement learning paradigms. The injected tokens function as conditioning context that \textbf{steers the policy's action distribution toward high-reward regions}. Unlike pure SFT, which suffers from distribution shift between data and policy rollouts~\citep{zhang2025onpolicyrlmeetsoffpolicy}, \inpaint{} maintains on-policy generation for the non-injected tokens, ensuring that gradient updates remain closer to the current policy's actual sampling distribution.

Finally, we augment \inpaint{} with additional reinforcement learning techniques that improve learning stability and performance, including entropy-based gradient filtering for inpainted tokens, and conduct comprehensive experiments across mathematical reasoning benchmarks. Through careful ablation analysis, we systematically evaluate each component of our approach to understand the mechanisms underlying inpainting-guided policy optimization. Our work makes the following key novel contributions:

\begin{itemize}[leftmargin=.18in]
    \item We propose IGPO, the \textbf{first work to utilize the unique inpainting capabilities of diffusion LLMs} for reinforcement learning. By strategically injecting partial reasoning traces during exploration, IGPO alleviates the inefficiency of sparse verifiable rewards and effectively mitigates the zero-advantage dilemma in group-based policy optimization methods such as GRPO. By substantially reducing the proportion of all-wrong groups (by $\sim$60\% as shown in Fig~\ref{fig:teaser} (b)) and preserving output diversity, our approach yields non-degenerate gradients.
    
    \item We propose a \emph{Length-Aligned} supervised fine-tuning strategy for full-attention based dLLMs using synthetically rewritten, concise reasoning traces. This design better aligns SFT data length with RL sampling and evaluation length, avoids the limitations of verbose traces, and provides stronger initialization for downstream RL. 
    
    \item Our full training recipe achieves substantial improvements on mathematical reasoning benchmarks, including \textbf{+4.9\% on GSM8K, +8.4\% on Math500, and +9.9\% on AMC} relative to the LLaDA-Instruct, achieving \textbf{state-of-the-art performance among full-attention based dLLMs} on these mathematical benchmarks.
    
    \item We conduct a comprehensive ablation study that disentangles the mechanisms of IGPO. We show that partial inpainting consistently outperforms full ground-truth inpainting by staying closer to the policy distribution in online RL, and propose an entropy-based gradient filtering mechanism that stabilizes training dynamics.
\end{itemize}

%% file: sections/background.tex
\section{Preliminaries}
\label{sec:background}
\subsection{Masked Diffusion Large Language Models}
Masked diffusion LLMs~\citep{austin2021structured, sahoo2024simple, shi2024simplified, ou2024your, loudiscrete} employ a forward diffusion process that progressively corrupts token sequences $x_0$ through introduction of $\mask$ tokens. This corruption process is parameterized by time $t \in [0, 1]$. At any given timestep $t$, the resulting sequence $x_t$ contains partial masking, where each token maintains a probability $\alpha_t$ of remaining unmasked. The noise schedule $\alpha_t$ exhibits strict monotonic decrease with respect to $t$. Complete masking occurs at $t=1$, where all tokens in $x_1$ become masked. The training procedure for masked dLLMs follows a forward process through definition of $\alpha_t$ and a bidirectional unmasking predictor $f_\theta$ with learnable parameters. During each training step, we stochastically sample timestep $t \in [0, 1)$ and apply token masking according to the designated forward process. Given these corrupted sequences, the training objective seeks to recover the original tokens. The standard optimization criterion employs the negative evidence lower bound (NELBO), which provides an upper bound for the negative log-likelihood (NLL) of the training data. For masked dLLMs, this NELBO reduces to a weighted NLL formulation, with weighting coefficients derived from transformations of $\alpha_t$~\citep[Equation (10)]{sahoo2024simple}.
For example, LLaDA~\citep{nie2025largelanguagediffusionmodels} specifies the forward process through $\alpha_t = 1 -t$, yielding the following NELBO formulation:
\begin{equation}
-  \mathbb{E}_
{{t\sim \mathcal{U}[0,1)}, \; x_0 \sim p_{\text{data}}, \; x_t \sim q_{t|0}(x_t | x_0)} \left[ \frac{1}{t} \sum_{k=1}^{|x_t|} \indicator[x_t^k = \mask] \log f_{\theta}(x_0^k \mid x_t) \right],
\label{eq:llada_loss}
\end{equation}
where $|x_t|$ denotes the sequence length of $x$, and $x^k$ represents the $k$-th token position.
The loss computation is restricted to tokens masked at timestep $t$. 

During prompt conditional generation, the model starts with a sequence where prompt tokens remain unmasked and continuation tokens are initially masked, then progressively unmasks the continuation tokens through ancestral sampling from the reverse process $p_\theta(x_s \mid x_t)$ for timesteps $t > s$, where the model $f_\theta$ provides the denoising predictions for masked positions. The reverse process maintains the property that unmasked tokens are carried over unchanged throughout all denoising steps.

\subsection{Policy Optimization for Masked Diffusion Large Language Models}
Policy‐gradient methods have gained widespread adoption for post-training LLMs~\citep{ouyang2022training, bai2022training, li2023remax, ahmadian2024back}. 
Online RL—particularly Group Relative Policy Optimization (GRPO)—has proved effective for improving language models~\citep{shao2024deepseekmath, guo2025deepseek, team2025kimi}. 
GRPO~\citep{shao2024deepseekmath} offers a computationally efficient alternative to PPO~\citep{schulman2017proximal} by using group-based statistics for advantage estimation, avoiding separate value-function training. 

The GRPO objective integrates clipping for stability and reverse KL regularization:
\begin{equation}
\label{eq:grpo_loss}
\resizebox{0.93\columnwidth}{!}{
$
\displaystyle
\mathcal{L}_{\text{GRPO}}(\theta)
= \mathbb{E}_{\stackrel{q \sim \mathcal{D}}{o_1, \ldots, o_G \sim \pi_{\theta_{\text{old}}}(\cdot | q)}}
    \left[ \frac{1}{G}\sum_{i=1}^G \frac{1}{|o_i|}\sum_{k=1}^{|o_i|} \min
    \left(\rho_i^k A_i, \text{clip}\left(\rho_i^k, 1-\varepsilon, 1 + \varepsilon \right) A_i\right)
- \beta D_{\text{KL}}\left[\pi_\theta(\cdot | q)\| \pi_{\text{ref}}(\cdot | q) \right]
\right],
$
}
\end{equation}
where $\rho_i^k = \frac{\pi_{\theta}(o_i^k|q, o_i^{<k})}{\pi_{\theta_{\text{old}}}(o_i^k|q, o_i^{<k})}$ is the likelihood ratio.

For a query $q$, GRPO samples $G$ responses $\{o_1,\ldots,o_G\}$ from the behavior policy $\pi_{\theta_{\text{old}}}$ and assigns a \emph{single sequence-level} advantage per response. 
Following \citet{liu2025understanding}, we use the unnormalized group-relative advantage
\begin{equation}
\label{eq:our_adv}
A_i \;=\; r(o_i) \;-\; \frac{1}{G}\sum_{j=1}^{G} r(o_j),
\end{equation}
where $r$ is the reward function.
This scalar $A_i$ is shared by \emph{all tokens} in $o_i$ when forming the tokenwise objective.

\paragraph{\textbf{Challenges in Applying GRPO to Diffusion LLMs}}
Applying GRPO to dLLMs is nontrivial. The objective in \cref{eq:grpo_loss} requires (i) \emph{token-level} probabilities for importance ratios and (ii) \emph{sequence-level} probabilities for KL regularization. 
Autoregressive models provide per-token conditionals via sequential factorization, enabling one-pass sequence scoring by the chain rule:
$\log \pi_{\mathrm{AR}}(o \mid q) = \sum_{k=1}^{|o|} \log \pi_{\mathrm{AR}}(o^k \mid q, o^{<k})$.
Accordingly, the reverse-KL decomposes as
\begin{equation}
D_{\mathrm{KL}}\!\left[\pi_\theta(\cdot \mid q)\,\big\|\,\pi_{\mathrm{ref}}(\cdot \mid q)\right]
= \mathbb{E}_{o \sim \pi_\theta(\cdot \mid q)} \left[
\sum_{k=1}^{|o|} \log \frac{\pi_\theta(o^k \mid q, o^{<k})}{\pi_{\mathrm{ref}}(o^k \mid q, o^{<k})}
\right].
\end{equation}
In contrast, dLLMs do not admit a sequential factorization of $\pi(o \mid q)$. dLLM's generation invokes the unmasking predictor $f_\theta$ across $M$ denoising steps, making $\pi_\theta$ a composition of $M$ mappings. Exact tokenwise probabilities would require marginalization over denoising trajectories and maintaining (and differentiating through) full denoising trajectories, which is prohibitive.

\paragraph{\textbf{Mean-Field Approximation for Efficient Optimization}}
To address this, recent work develops efficient approximations for policy optimization in masked diffusion LLMs.
DiffuGRPO~\citep{zhao2025d1} employs a mean-field approximation that yields \emph{single-pass} estimates of both token-level and sequence-level terms, replacing explicit multi-step unrolling with a single-sample Monte Carlo estimate. 
While this introduces bias relative to the exact diffusion policy, it provides a practical framework for GRPO-style optimization on dLLMs.
In our method, we adopt the mean-field estimators of \citet{zhao2025d1} to compute the token-level importance ratios $\rho_i^k$ and the reverse-KL term with one forward pass per policy.

%% file: sections/method.tex
\begin{figure}[t]
    \centering
    \includegraphics[width=\columnwidth]{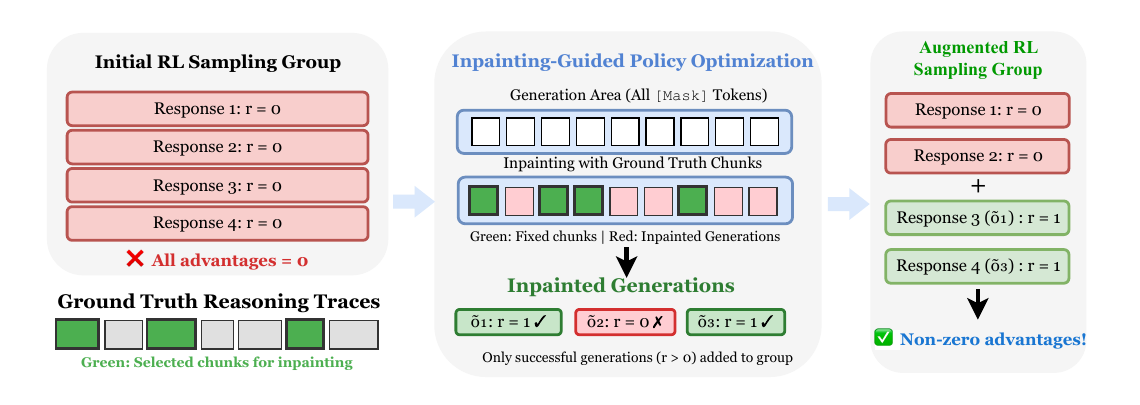}
     \caption{\textbf{Overview of \inpaint{}:} When all sampled responses yield identical incorrect rewards (zero-advantage scenario), we perform hint-guided inpainting by generating additional responses using ground truth reasoning chunks as injected hints. Ground truth traces $y^*$ are segmented into variable-length chunks, and selected chunks are injected as fixed hints during generation while the model generates the remaining tokens. We then replace a fraction of the original incorrect responses with correct responses generated through inpainting, creating reward variance that enables non-zero advantages for effective policy gradient updates.}
\end{figure}
\section{Methods}
\subsection{\inpaint{}: Inpainting Guided Policy Optimization}
\label{sec:method_igpo}

\paragraph{\textbf{Zero-Advantage Dilemma.}}
In the GRPO framework, when sampling $G$ responses $\{o_1, o_2, \ldots, o_G\}$ for a given prompt $q$, the advantage computation in \cref{eq:our_adv} relies on reward variance across the group. However, when all responses receive identical rewards---either all correct or all incorrect ---the advantages become zero:
$
A_i^k = r(o_i) - \frac{1}{G}\sum_{j=1}^{G} r(o_j) = 0.
$ This zero-advantage scenario renders the policy gradient component degenerate. Specifically, the clipped surrogate objective collapses to zero regardless of whether the update lies in the clipped or unclipped region, since both terms contain $A_i^k=0$. The policy gradient therefore becomes:

\begin{equation}
\nabla_\theta \mathcal{L}(\theta) =
\mathbb{E}\!\left[ \frac{1}{G}\sum_{i=1}^G \frac{1}{|o_i|}\sum_{k=1}^{|o_i|}
A_i \,\rho_i^k \,\nabla_\theta \log \pi_\theta(o_i^k \mid q) \right] = 0,
\end{equation}

As a result, no meaningful policy update can be extracted from the reward signal, wasting compute sampling these responses. \textbf{In this work, we specifically focus on mitigating the \emph{all-wrong} case.}

\paragraph{\textbf{Masked dLLM Generation and Inpainting.}}
In full-attention masked dLLM generation, the model input at denoising step 0 is the concatenation $[q; z_{\mask}]$, where $q$ represents the prompt and $z_{\mask} = [\mask, \mask, \ldots, \mask]$ denotes a fully masked completion sequence of predetermined length $L$. The generation process progressively unmasks these positions through iterative denoising until producing the final output.

\emph{Hint injection} modifies this formulation by fixing selected positions of $z_{\mask}$ to ground-truth tokens. Formally, given a ground-truth reasoning trace $y^* = [y^*_1, y^*_2, \ldots, y^*_{|y^*|}]$ and a binary mask $m \in \{0,1\}^L$ indicating which positions to inject as fixed hints, we construct the hint-injected initialization:
\begin{equation}
z^{\text{hint}}[i] =
\begin{cases}
y^*[i] & \text{if } m[i] = 1 \text{ and } i \leq |y^*|, \\
\mask & \text{otherwise}.
\end{cases}
\end{equation}
The masked dLLM then performs bidirectional denoising on $[q; z^{\text{hint}}]$ through the inpainting process, leveraging both the prompt and injected hint tokens to generate coherent responses. The injected hint tokens remain fixed throughout the iterative denoising steps.

\paragraph{\textbf{Constructing Hint Patterns for Inpainting.}}
To construct meaningful hint patterns for the inpainting process, we segment the ground truth reasoning trace $y^*$ into variable-length contiguous chunks $\mathcal{C} = \{c_1, c_2, \ldots, c_N\}$, where each chunk length $|c_j|$ is sampled from $\mathcal{U}[s_{\text{min}}, s_{\text{max}}]$. We explicitly exclude the final answer tokens from chunking to prevent reward hacking behaviors where the model ignores reasoning and collapses. For a given hint injection ratio $\eta \in [0, 1]$, we randomly select $\lfloor \eta \cdot N \rfloor$ chunks and set their corresponding positions in the binary mask $m$ to 1 for hint injection.

\input{sections/algo_box}

\paragraph{\textbf{Elastic Inpainting-Triggered Sampling.}}
With the above inpainting setup, we design IGPO (as in \cref{algo:igpo}) to be \textbf{elastic}: hint injection is only triggered when all sampled responses in a group yield incorrect rewards (the zero-advantage case), and when activated, both the hint injection ratio and chunk sizes are randomized to provide diverse training signals. When detecting that all sampled responses $\{o_1, \ldots, o_G\}$ for query $q$ yield identical correctness rewards $r(o_i) = 0$, we generate an additional set of responses $\{\tilde{o}_1, \ldots, \tilde{o}_G\}$ through the inpainting process. Each response $\tilde{o}_i$ is generated via inpainting with a distinct hint injection ratio $\eta_i \sim \mathcal{U}[\eta_{\text{low}}, \eta_{\text{high}}]$ to ensure diverse hint densities. Following inpainting generation, we evaluate the correctness of $\{\tilde{o}_i\}$ and only use the correct ones for replacement. Specifically, we replace $K = \min(|\{\tilde{o}_i : r(\tilde{o}_i) = 1\}|, \lfloor \lambda G \rfloor)$ of the original incorrect responses with correct responses generated through inpainting, where $\lambda \in (0, 1)$ controls the replacement fraction to maintain non-zero advantage variance.

The complete IGPO objective modifies the GRPO formulation by incorporating the augmented sampling procedure:
\begin{equation}
\label{eq:igpo_loss}
\resizebox{0.95\columnwidth}{!}{
$
\displaystyle
\mathcal{L}_{\text{IGPO}}(\theta)
= \mathbb{E}_{\stackrel{q \sim \D}{\textcolor{blue}{\{o_1, \ldots, o_{G-K}, \tilde{o}_1, \ldots, \tilde{o}_K\} \sim \textbf{\text{IGPO-Sample}}(\pi_\theta, q, y^* )}}}
    \left[ \left( \frac{1}{G}\sum_{i=1}^G \frac{1}{L_i}\sum_{k=1}^{L_i} \min
    \left(\rho_i^k A_i^k, \text{clip}\left(\rho_i^k, 1-\varepsilon, 1 + \varepsilon \right) A_i^k\right) \right)
- \beta D_{\text{KL}}\left[\pi_\theta(\cdot | q)\| \pi_\text{ref}(\cdot | q) \right]
\right],
$
}
\end{equation}
where $\text{IGPO-Sample}(\pi_\theta, q, y^*)$ denotes the augmented sampling procedure that applies inpainting-based augmentation when zero-advantage scenarios are detected, producing the augmented RL sampling group $\{o_1, \ldots, o_{G-K}, \tilde{o}_1, \ldots, \tilde{o}_K\}$ containing $(G-K)$ original responses and $K$ verified correct inpainted responses $\{\tilde{o}_i\}$ after replacement, where $L_i$ denotes the length of the $i$-th response (whether $o_i$ or $\tilde{o}_i$). Crucially, only inpainted responses that pass correctness verification are included in the augmented group, satisfying $r(\tilde{o}_i) = 1$. Advantages $A_i^k$ are computed normally according to \cref{eq:our_adv}. We built \inpaint{} with DiffuGRPO~\citep{zhao2025d1}'s log probability estimation methods, where all completion tokens are masked during estimation and we remove the random masking applied to prompt tokens as done in DiffuGRPO. Since we use a small number of policy iterations (i.e. 4), this alleviates the need for random prompt masking to reduce overfitting. Inspired by~\citet{zheng2025group}, we compute sequence-level importance-ratio through mean-field approximation for stability purposes.

\paragraph{\textbf{Entropy-based Gradient Filtering for Hint Tokens.}}
When applying \inpaint{} to zero-advantage scenarios, the responses generated through inpainting contain ground truth reasoning chunks that originate from a different distribution than the current policy $\pi_\theta$. This creates an off-policy learning scenario where gradient updates from ground truth tokens can conflict with the model's current beliefs, particularly at positions where the model has high confidence (low entropy). To mitigate potential training instability from this distribution mismatch, we implement an entropy-based filtering approach that restricts learning to hint token positions where the model exhibits sufficient uncertainty, as inspired by \citet{huang2025blendingsupervisedreinforcementfinetuning}. Specifically, for each hint token position (i.e., positions with injected ground-truth tokens) we compute the entropy. We then apply gradient updates only to the top $\tau$ percentile of hint token positions with highest entropy values. This selective learning strategy serves two purposes: high-entropy positions represent genuine decision boundaries where the model is naturally uncertain and thus more receptive to external guidance, and they correspond to flatter probability distributions that yield more stable gradient updates when incorporating ground truth information. This approach controls the policy shift by focusing learning on positions where the model is already open to change, rather than forcing updates against strong existing beliefs at low-entropy positions.


\subsection{Length-Aligned SFT via Concise Reasoning Trace Rewriting}
\label{sec:concise_reasoning_sft}

We seek a stronger RL initialization via SFT that matches the model's generation length across SFT, RL sampling, and evaluation time. Full-attention masked dLLMs like LLaDA do not utilize KV cache optimization by default~\citep{wu2025fastdllmtrainingfreeaccelerationdiffusion}, requiring complete sequence attention computation at each denoising step. This computational constraint dominates online RL training time. Consequently, we are constrained to limit RL generation lengths to 256 tokens for computational feasibility and faster convergence with a reduced exploration space. Also, most evaluation setups for LLaDA in recent works~\citep{zhao2025d1, li2025llada, nie2025largelanguagediffusionmodels} use response sequences with fixed lengths up to 1024 tokens. However, popular reasoning SFT corpora (e.g., OpenR1) contain verbose reasoning traces that frequently exceed 10k tokens. SFT on these corpora introduces distribution mismatch across SFT training, RL and evaluation sampling. Moreover, these reasoning datasets contain many repeated reflective behaviors unsuitable for limited context generation.

\paragraph{\textbf{Addressing Generation Length Mismatch Through Concise Reasoning Trace Rewriting.}} To address this generation length mismatch, we propose systematic rewriting of verbose reasoning traces into concise, structured versions that preserve essential logical flow while conforming to the computational constraints of full-attention masked dLLMs. We employ LLaMA-4-Maverick~\citep{meta2025llama} (with designed prompts in \cref{appendix:revision_prompt}) to eliminate redundant reflections, condense multi-sentence elaborations into precise, mathematically rigorous statements, and maintain essential reasoning steps. Our \emph{Length-Aligned SFT} trains LLaDA exclusively on these rewritten traces, providing better initialization for RL learning by eliminating the need for implicit length compression during RL and allowing the model to focus on reasoning quality improvements within fixed computational bounds. Empirical evaluation demonstrates superior performance compared to training on original verbose traces. Additionally, we find that masked dLLMs benefit from longer training epochs (e.g. 200) compared to AR LLMs, as has also been noted in recent works~\citep{ni2025difflm, prabhudesai2025diffusionbeatsautoregressivedataconstrained}.

%% file: sections/algo_box.tex
\begin{algorithm}[t]
\caption{\inpaint{}: Inpainting-Guided Policy Optimization for Masked dLLMs}
\begin{algorithmic}[1]
\Require Reference model $\pi_{\text{ref}}$, prompt distribution $\mathcal{D}$, ground-truth reasoning traces $\{y^*\}$, number of completions per prompt $G$, number of inner updates $\mu$, hint injection ratio range $[\eta_{\text{low}}, \eta_{\text{high}}]$, replacement fraction $\lambda$, entropy filter threshold $\tau$, positive boost coefficient $\alpha$, chunk size range $[s_{\text{min}}, s_{\text{max}}]$
\State Initialize $\pi_{\theta} \gets \pi_{\text{ref}}$
\While{not converged}
    \State $\pi_{\text{old}} \gets \pi_{\theta}$
    \State Sample prompt $q \sim \mathcal{D}$ and $G$ responses $o_i \sim \pi_{\text{old}}(\cdot \mid q), \; i \in [G]$ and compute rewards $r_i$ 
    \If{all $r_i = 0$ (zero-advantage case)}
        \State Segment ground-truth reasoning $y^*$ into chunks $\{c_1,\ldots,c_N\}$ with $|c_j| \sim \mathcal{U}[s_{\text{min}}, s_{\text{max}}]$
        \For{$i = 1,\ldots,G$}
            \State Sample hint injection ratio $\eta \sim \mathcal{U}[\eta_{\text{low}}, \eta_{\text{high}}]$
           and select $\lfloor \eta N \rfloor$ chunks from $\{c_1,\ldots,c_N\}$ randomly
            \State Inject selected chunk tokens as fixed hints at corresponding positions
            \State Generate $\tilde{o}_i$ via inpainting: iteratively denoise the remaining masked positions while keeping hint tokens fixed
        \EndFor
        \State Evaluate rewards $r(\tilde{o}_i)$ and replace up to $\lfloor \lambda G \rfloor$ incorrect responses with correct responses generated through inpainting
    \EndIf
    \State Compute advantages $A_i^k$ on the updated response set using Eq.~\ref{eq:our_adv}
    \For{$n=1,\ldots,\mu$}
        \State Estimate log-probabilities under $\pi_\theta, \pi_{\text{old}}, \pi_{\text{ref}}$
        \State For hint token positions, update only top-$\tau$ percentile highest-entropy positions 
        \State Update $\pi_\theta$ via $\mathcal{L}_{\text{IGPO}}(\theta)$ (Eq.~\ref{eq:igpo_loss})
    \EndFor
\EndWhile
\State \Return $\pi_{\theta}$
\end{algorithmic}
\label{algo:igpo}
\end{algorithm}

%% file: sections/experiments.tex
\section{Experiments}
\vspace{-2mm}
To investigate how the inpainting capabilities of masked dLLMs can address the exploration challenges in RL, we conduct comprehensive experiments to answer the following main research questions:

\begin{itemize}[leftmargin=15pt,topsep=0pt,itemsep=0.3pt,parsep=0pt]
    \item[(1)] How effectively does our complete training approach (Length-aligned SFT with rewritten reasoning traces followed by reinforcement learning with \inpaint{}) improve the mathematical reasoning performance of LLaDA and reduce all-wrong groups occurrences? (\S\ref{sec:main_results})
    \item[(2)] How does partial hint injection in \inpaint{} bridge on-policy generation with ground truth guidance, and how does this improve learning compared to full supervision? (\S\ref{sec:ablations})
    \item[(3)] How do other key design choices—including entropy filtering thresholds and reasoning trace rewriting—affect RL training dynamics and learning stability? (\S\ref{sec:ablations})
\end{itemize}

\subsection{Complete Training Recipe}
\label{sec:training_pipeline}

Our complete learning framework consists of a two-stage pipeline:

\textbf{Stage 1: Supervised Fine-Tuning with Rewritten Traces.} We begin with \emph{Length-Aligned} SFT on the LLaDA-8B-Instruct model using the OpenR1-Math-220K dataset's default split (94k math problems), but with all reasoning traces rewritten (See \cref{appendix:sft_revision_token_length} for length distribution before and after revision). This ensures consistency between training distribution and downstream RL/evaluation phases by aligning trace lengths.

\textbf{Stage 2: Reinforcement Learning with \inpaint{}.} Following \emph{Length-aligned} SFT, we apply \inpaint{} to further enhance reasoning capabilities through strategic inpainting-guided policy optimization. We utilize the reasoning traces from the MetaMathQA dataset for the elastic inpainting process, creating effective guidance signals that fit within our computational constraints. Detailed hyperparameters are provided in \cref{appendix:hyperparameters}.

\subsection{Experimental Setup}

We conduct experiments using LLaDA-8B-Instruct as the base model with a sampling temperature of 1.2 for RL online generation. For reinforcement learning, we train on the MetaMathQA dataset, specifically using the ``Answer Augmentation" split and combining questions from both GSM8K and MATH500. After deduplicating identical questions, we obtain 12,794 unique training examples. For supervised fine-tuning, we utilize the OpenR1-Math-220K dataset with rewritten reasoning traces as described in Section~\ref{sec:concise_reasoning_sft}. We evaluate our approach on three mathematics benchmarks: GSM8K, MATH500 and AMC. Experiments are conducted on 8×8 80GB H100 GPUs. For UniGRPO~\citep{yang2025mmada} baseline, we reproduce based on their Algorithm 1. We provide detailed experiment hyperparameter setups in \cref{appendix:hyperparameters}.

\vspace{-2mm}

\subsection{Main Results}
\label{sec:main_results}
\begin{figure}[h]
    \centering
    \begin{subfigure}[b]{0.4\textwidth}
        \centering
        \includegraphics[width=\textwidth]{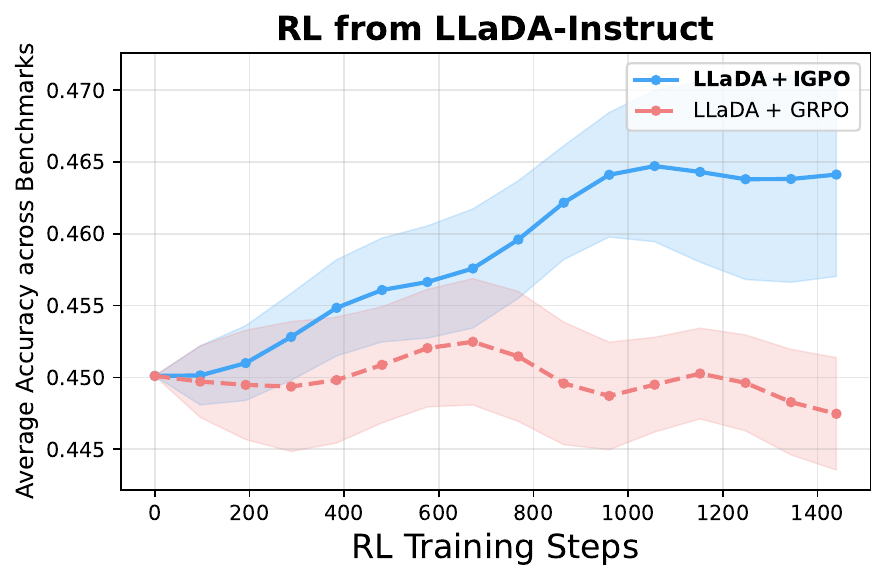}
        \label{fig:igpo_vs_grpo}
    \end{subfigure}
 \hspace{5mm} 
    \begin{subfigure}[b]{0.4\textwidth}
        \centering
        \includegraphics[width=\textwidth]{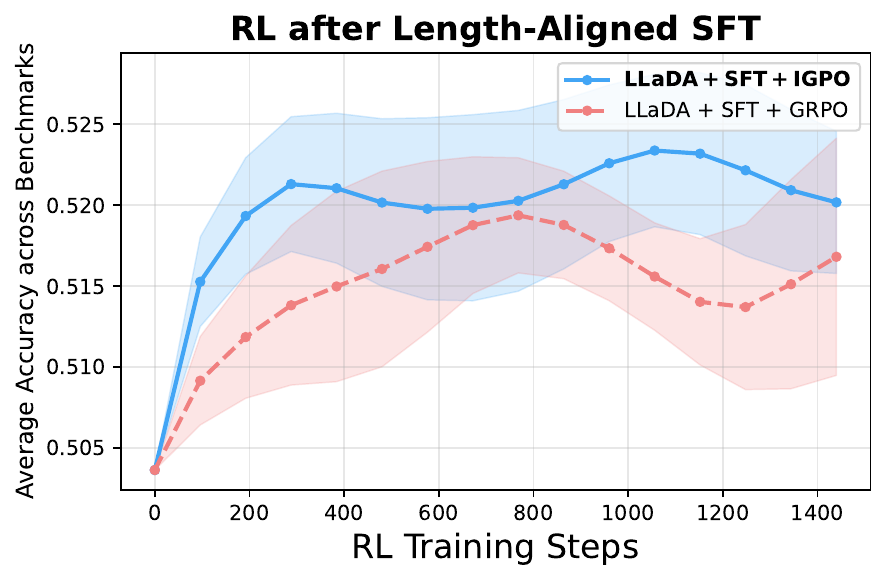}
        \label{fig:length_aligned_sft}
    \end{subfigure}
    \vspace{-4mm}
\caption{\textbf{RL training curves of \inpaint{} versus normal GRPO sampling.} (a) Starting from LLaDA-8B-Instruct. (b) Starting from the \emph{length-aligned} SFT checkpoint. \inpaint{} exhibits superior and more stable training performance under both initialization checkpoints. Results are averaged over 3 random seeds across three mathematical reasoning benchmarks (GSM8K, MATH500, and AMC), with standard errors shown as shaded regions.}
    \label{fig:mainresult_combined_training_dynamics}
\end{figure}
\begin{table}[h]
    \centering
    \small
\caption{Performance across multiple mathematics tasks. GSM8K and MATH500 are evaluated with pass@1 at temperature of 0.0, and AMC with avg@16 at temperature 0.1.
    Underlined scores indicate the best \emph{within each initialization group}.
    Parenthesized deltas typeset via \pos{} denote absolute percentage-point improvements \emph{relative to the LLaDA-8B-Instruct} baseline.}
    
    \renewcommand{\arraystretch}{1.0}
    \setlength{\tabcolsep}{8pt}
    \scalebox{0.92}{ 
    \begin{tabular}{@{}l
        S[table-format=2.1]@{}l
        S[table-format=2.1]@{}l
        S[table-format=2.1]@{}l
        S[table-format=2.1]@{}l@{}}
        \toprule
        \textbf{Model}
            & \multicolumn{2}{c}{\shortstack{\textbf{GSM8K}\\(pass@1)}}
            & \multicolumn{2}{c}{\shortstack{\textbf{MATH500}\\(pass@1)}}
            & \multicolumn{2}{c}{\shortstack{\textbf{AMC}\\(avg@16)}}
            & \multicolumn{2}{c}{\shortstack{\textbf{Average}\\{}}} \\
         \midrule
        \multicolumn{9}{c}{\textit{Similar-sized autoregressive LLMs}} \\
        \midrule
          LLaMA3-8B~\citep{llama3modelcard}     & 79.6 &  & 30.0 &  & 
          \multicolumn{2}{l}{--} &
          \multicolumn{2}{l}{--} \\
      Qwen2.5-7B~\citep{qwen2.5}    & 85.4 &  & 49.8 &  & 
          \multicolumn{2}{l}{--} &
          \multicolumn{2}{l}{--} \\
        \midrule
        \multicolumn{9}{c}{\textit{Prior masked dLLM baselines}} \\
        \midrule
      
        Dream-7B \citep{dream2025}       & 77.2 &  & 39.6 &  & \multicolumn{2}{l}{--} & \multicolumn{2}{l}{--} \\
        d1-LLaDA \citep{zhao2025d1}      & 82.1 &  & 40.2 &  & \multicolumn{2}{l}{--} & \multicolumn{2}{l}{--} \\
        wd1 \citep{tang2025wd1weightedpolicyoptimization} & 82.3 &  & 39.0 &  & \multicolumn{2}{l}{--} & \multicolumn{2}{l}{--} \\
        LLaDA-1.5 \citep{li2025llada}    & 83.3 &  & 42.6 &  & 13.6 &  & 46.5 & \\
  LLaDA-Instruct \citep{nie2025largelanguagediffusionmodels}
  & 81.5 & {\posbaseline{0}}
  & 39.0 & {\posbaseline{0}}
  & 14.5 & {\posbaseline{0}}
  & 45.0 & {\posbaseline{0}} \\
        \midrule
        \multicolumn{9}{c}{\textit{RL from LLaDA-Instruct}} \\
        \midrule
       LLaDA-Instruct + UniGRPO~\citep{yang2025mmada}                & 82.2 & \pos{0.7} & 39.2 & \pos{0.2} & 15.0 & \pos{0.5} & 45.5 & \pos{0.5} \\
        LLaDA-Instruct + DiffuGRPO~\citep{zhao2025d1}               & 81.9 & \pos{0.4} & 40.2 & \pos{1.2} & 17.5 & \pos{3.0} & 46.5 & \pos{1.5} \\
        LLaDA-Instruct + \inpaint{}  (ours)
        & \underline{83.6} & \pos{2.1} & \underline{42.8} & \pos{3.8} & \underline{18.1} & \pos{3.8} & \underline{48.2} & \pos{3.2} \\
        \midrule
        \multicolumn{9}{c}{\textit{Length-aligned SFT on LLaDA-Instruct and RL on SFT checkpoint}} \\
        \midrule
       LLaDA-Instruct + Length-aligned SFT (ours)                       & 83.6 & \pos{2.1} & 45.2 & \pos{6.2} & 22.3 & \pos{7.8} & 50.4 & \pos{5.4} \\
       LLaDA-Instruct + Length-aligned SFT + \inpaint{}   (ours)        & \underline{\bfseries 86.4} & \pos{4.9} & \underline{\bfseries 47.4} & \pos{8.4} & \underline{\bfseries 24.4} & \pos{9.9} & \underline{\bfseries 52.7} & \pos{7.7} \\
        \bottomrule
    \end{tabular}
    }
    \label{tab:main_results}
\end{table}

\newpage

As shown in \cref{tab:main_results}, our training recipe demonstrates consistent improvements across all mathematical reasoning benchmarks. With \emph{Length-Aligned} SFT on rewritten traces, LLaDA improves by 2.1\% on GSM8K and 6.2\% on MATH500 compared to the base LLaDA-8B-Instruct model. When applying \inpaint{} on top of the SFT model, we observe further improvements of 2.8\% on GSM8K to 86.4\% and 2.2\% on MATH500 to 47.4\%. The complete two-stage pipeline yields cumulative improvements of 4.9\% on GSM8K and 8.4\% on MATH500 relative to the LLaDA-Instruct baseline. On the challenging AMC benchmark, our approach achieves 24.4\% (avg@16), representing a 9.9\% improvement over the baseline.

As shown in \cref{fig:mainresult_combined_training_dynamics}, \inpaint{} exhibits more stable training dynamics compared to standard GRPO sampling when initializing from either the base LLaDA checkpoint or after SFT. \inpaint{} effectively reduces the all-wrong group ratio by around 60\%, as shown in \cref{fig:teaser}(b). Our final model (LLaDA + SFT + \inpaint{}) outperforms all baseline approaches including the recent LLaDA-1.5 model across all evaluated benchmarks. Notably, even without SFT, applying \inpaint{} directly on LLaDA achieves better performance than the previous LLaDA-1.5 and other RL methods for full-attention dLLMs, establishing a new state-of-the-art recipe for mathematical reasoning in masked diffusion language models.

\subsection{Analysis and Ablation Studies}
\label{sec:ablations}
\begin{wrapfigure}{r}{0.5\textwidth}
    \centering
    \includegraphics[width=0.48\textwidth]{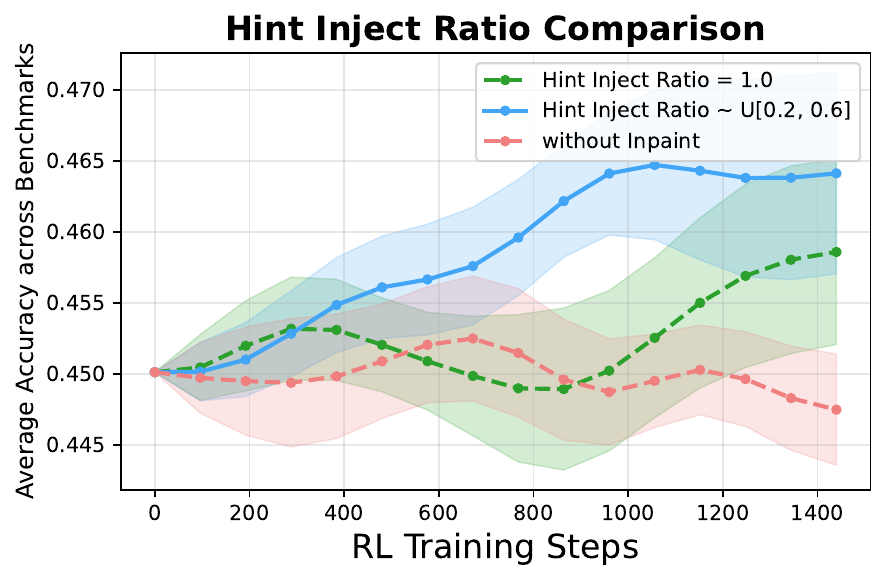}
    \caption{\textbf{Impact of hint injection ratio on performance} across 3 datasets, averaged over 3 seeds with standard error shown as shaded areas. We compare partial hint injection ($\eta \sim \mathcal{U}[0.2, 0.6]$) versus full hint injection ($\eta = 1.0$). Partial hint injection consistently outperforms full hint injection, demonstrating the benefits of self-generated reasoning. Both hint-guided inpainting variants outperform the baseline without any hint injection.}
    \label{fig:inpainting-ratio-comparison}
\end{wrapfigure}

\vspace{4mm}
\paragraph{\textbf{Self-generated inpainted traces provide better learning signal than ground truth traces.}}

The results in \cref{fig:inpainting-ratio-comparison} show that partial hint injection achieves higher performance than full hint injection. When the hint injection ratio varies within the lower range, the model needs to generate self-rationalized inpainting traces (with an example shown in \cref{appendix:inpainted_examples}), and only those that lead to correct solutions are added to the group for gradient updates. Through inpainting, the model attempts to coherently connect provided hint chunks with its own reasoning steps. The inpainted generation produces a learning signal that bridges the gap between the model's current capabilities and the target behavior. The self-generated portions reflect the model's current reasoning patterns and are more "on-policy" while incorporating structural guidance from ground truth chunks, resulting in more effective policy optimization compared to pure supervised learning, reducing the distributional mismatch. \textbf{This bridging of SFT and online RL through partial self-generation leads to more effective policy optimization.}

\begin{figure}[h]
    \centering
    \includegraphics[width=0.7\textwidth]{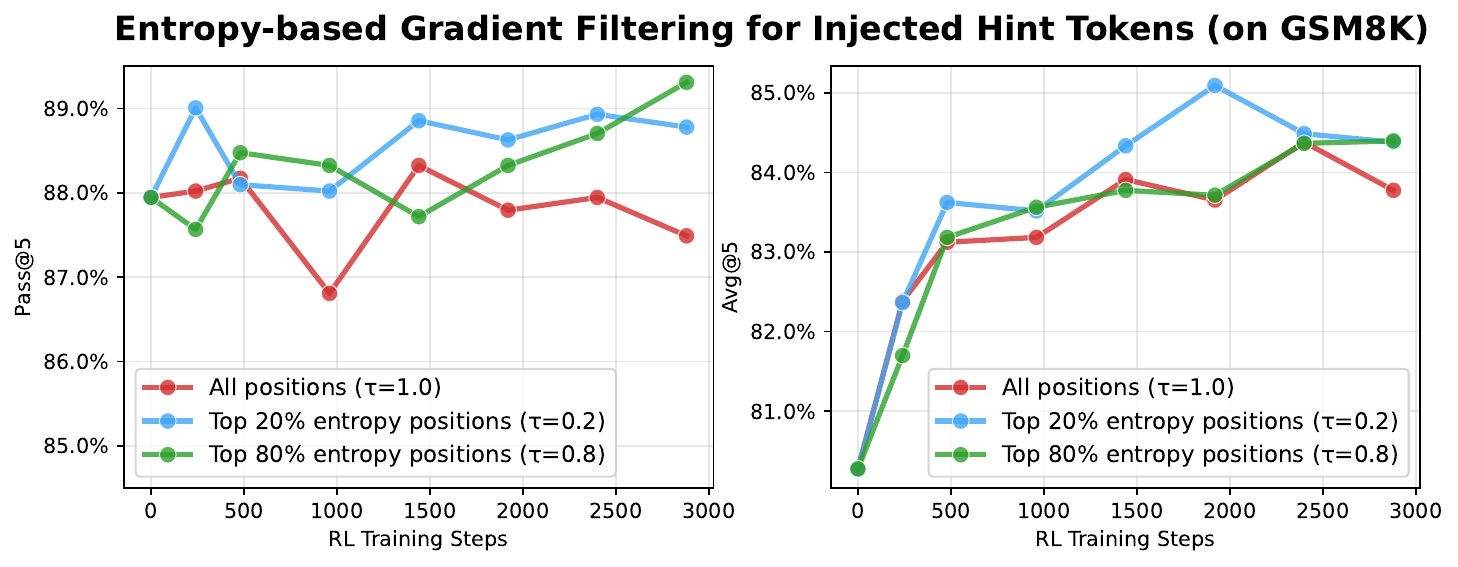}
\caption{\textbf{Impact of entropy clipping threshold on hint tokens.} Performance comparison across different entropy clipping thresholds $\tau$ applied to hint token positions in \inpaint{}, where $\tau=0.2$ represents learning from only the top 20\% highest-entropy hint token positions, while $\tau=1.0$ indicates learning from all hint token positions without filtering. This results is run on
GSM8K with temperature of 0.1 and generation length of 256.}
    \label{fig:entropy_clipping_analysis}
\end{figure}

\paragraph{\textbf{Entropy clipping prevents training instability from off-policy tokens.}}
As shown in Figure~\ref{fig:entropy_clipping_analysis}, we observe that learning from only the top 20\% highest-entropy hint token positions ($\tau=0.2$) achieves the best performance and exhibits the most stable training dynamics. In contrast, learning from all hint token positions ($\tau=1.0$) or a large fraction ($\tau=0.8$) leads to more unstable training with performance fluctuations compared to lower values like 0.2. This empirical finding supports our motivation that restricting gradient updates to high-entropy positions prevents the destabilizing effects of large gradients on high-entropy positions. The validates the necessity of entropy-based filtering when incorporating ground truth traces from hint-guided inpainting into policy gradient training.

\newpage
\paragraph{\textbf{Effect of reasoning trace rewriting for SFT and subsequent RL training.}  }

\begin{figure}[h]
\centering
\includegraphics[width=0.7\textwidth]{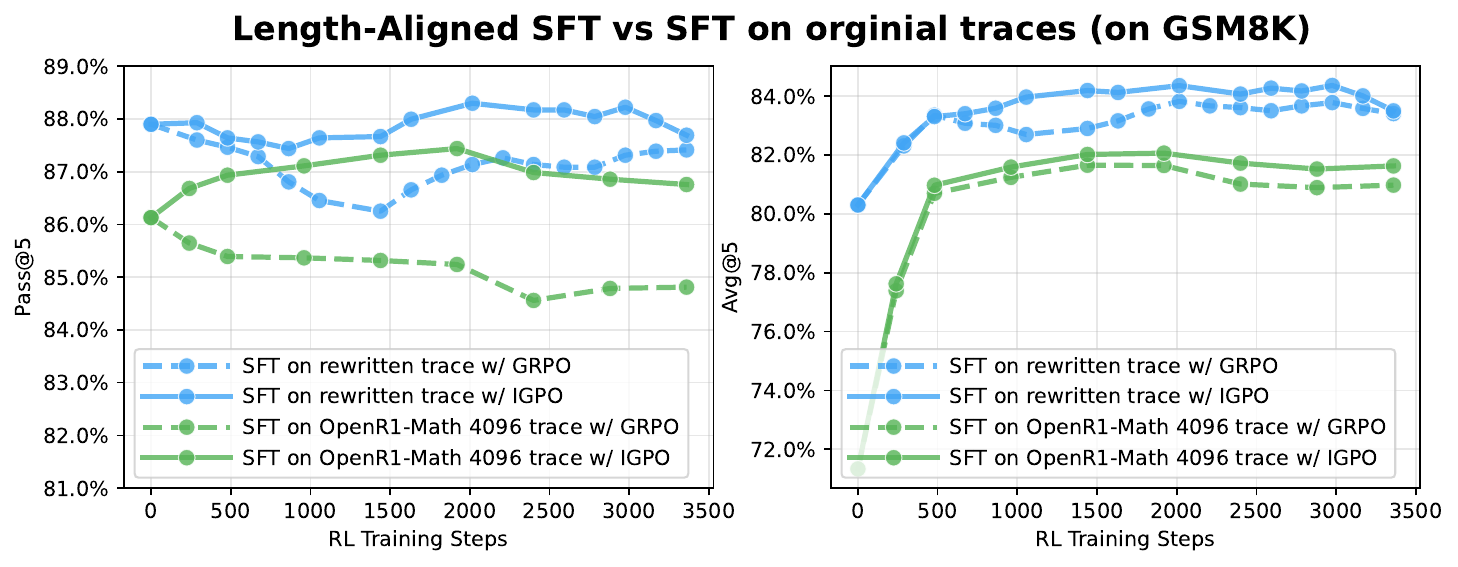}
\caption{\textbf{SFT and RL training dynamics with rewritten versus original traces.} 
We compare two settings: (1) models fine-tuned with SFT on our rewritten concise traces (maximum length 1024 tokens), and (2) models fine-tuned with SFT on the original OpenR1-Math traces truncated at LLaDA’s 4096-token context limit. 
After SFT, both checkpoints are used as initialization for RL training (either standard RL or \inpaint{}). 
The rewritten traces yield higher SFT performance and lead to superior final RL accuracy. 
Across both initializations, \inpaint{} consistently outperforms standard RL, maintaining stable pass@5 performance while standard RL suffers from declining diversity. This results is run on GSM8K with temperature of 0.1 and generation length of 256.}
\label{fig:sft_rl_rewritten_vs_original}
\end{figure}

The results in \cref{fig:sft_rl_rewritten_vs_original} illustrate two key findings. 
First, SFT on rewritten reasoning traces produces substantially stronger checkpoints than SFT on the original traces. 
Our rewritten traces eliminate verbose reflection behaviors and compress reasoning into concise trajectories (up to 1024 tokens), which are better aligned with LLaDA’s generation budget (256 tokens) and evaluation sequence length. 
This alignment improves SFT accuracy at step 0 relative to models trained on the longer 4096-token traces.  

Second, while RL training can partially compensate for weaker SFT checkpoints---the models trained on 4096-token traces recover accuracy rapidly in early RL steps---starting from stronger rewritten SFT checkpoints leads to consistently higher final performance. 
Importantly, across both initialization settings, \inpaint{} outperforms standard RL without inpainting: \inpaint{} preserves output diversity and stabilizes pass@5 performance throughout training, whereas standard RL exhibits a decline in pass@k metrics, indicative of reduced exploration and mode collapse.  

%% file: sections/related_works.tex
\section{Related Work}

\subsection{Diffusion Language Models}
Diffusion language models was first explored through continuous approaches that map discrete text to continuous representations, including learned embeddings, sequence-to-sequence conditioning, and binary bit representations \citep{Chen2022AnalogBG, Li-2022-DiffusionLM, gong2022diffuseq}. Recently, discrete diffusion language models have been scaled up significantly, with masked diffusion established as a specific instance of discrete diffusion \citep{austin2021structured, sahoo2024simple, shi2024simplified, ou2024your, nie2024scaling}. Notable developments include DiffuLLaMA \citep{gong2025scaling} and Dream \citep{dream2025}, both adapted from pretrained autoregressive LLMs. LLaDA \citep{nie2025largelanguagediffusionmodels} represents a breakthrough as a masked diffusion LLM trained from scratch using full-attention,  achieving performance comparable to similarly-sized autoregressive models. These approaches are predominantly based on masked modeling. Unlike these full-attention dLLMs, Block Diffusion \citep{arriola2025block} introduced a hybrid approach that models sequences block-by-block while applying diffusion within each block, enabling flexible length generation and improved inference efficiency through kv-caching. Recent commercial models like Mercury \citep{inception2025mercury} and Gemini Diffusion~\citep{deepmind2025gemini} have demonstrated the practical viability of diffusion-based code generation, achieving performance comparable to leading autoregressive models while offering significantly faster inference. More recent works have introduced caching and parallel decoding algorithms \citep{wu2025fastdllmtrainingfreeaccelerationdiffusion, liu2025dllm, ma2025dkvcachecachediffusionlanguage, israel2025acceleratingdiffusionllmsadaptive, sahoo2025esotericlanguagemodels,hu2025acceleratingdiffusionlanguagemodel} that significantly improve inference efficiency for masked diffusion language models. In this work, we focus on full-attention masked dLLMs.

\subsection{Reinforcement Learning for Diffusion Language Models}
Applying reinforcement learning to diffusion language models presents unique challenges compared to autoregressive models. The primary obstacle is the intractability of likelihood functions in diffusion models, which necessitates approximating sequence-level likelihoods for policy optimization. This requirement introduces computational overhead and potential bias, particularly when approximation errors occur in policy ratios used for importance sampling. d1 proposed diffu-GRPO~\citep{zhao2025d1} which adopts an efficient approximation through mean-field approximation. MMaDA~\citep{yang2025mmada} and diffucoder's coupled-GRPO~\citep{gong2025diffucoder} further improve the masking strategy in log probabilities estimation to achieve better learning efficiency.  LLaDA 1.5~\citep{li2025llada} tackles the variance issues in ELBO-based likelihood estimates through preference optimization. Recently, wd1~\citep{tang2025wd1weightedpolicyoptimization} addresses these challenges by reformulating policy optimization as a weighted likelihood objective that eliminates the need for policy ratios. SDPO~\citep{han2025discretediffusiontrajectoryalignment} decomposes the diffusion trajectory alignment problem into stepwise subproblems that align the posterior at each diffusion step. Our inpainting method can also be applicable to some of the above online RL methods. A closely related work in RL for AR LLMs is Prefix-RFT~\citep{huang2025blendingsupervisedreinforcementfinetuning}, which samples prefixes from demonstrations to guide online exploration, though this is limited to left-to-right sequential generation that does not leverage the bidirectional conditioning capabilities of diffusion LLMs.

%% file: sections/conclusion.tex
\section{Conclusion}
In this work we introduce \inpaint{}, a novel reinforcement learning algorithm that utilizes the inpainting capabilities of masked diffusion language models. By strategically incorporating ground truth reasoning hints during the denoising process, \inpaint{} steers the policy's action distribution toward high-reward regions and mitigates the fundamental exploration challenge in RL. \inpaint{} addresses the zero-advantage dilemma by creating  reward variance that enables effective policy gradient updates when standard sampling procedures yield uniform outcomes. As part of our recipe, \emph{Length-Aligned} SFT mitigates the length distribution gap between SFT training, RL, and evaluation sampling, and provides a stronger initialization for downstream RL. Our comprehensive evaluation demonstrates that this methodology, combined with the stabilization mechanism of entropy-based gradient filtering, establishes new state-of-the-art performance among full-attention masked dLLMs across multiple mathematical reasoning benchmarks. This contribution presents a novel paradigm for reinforcement learning in masked diffusion language models, demonstrating how architectural properties can be systematically leveraged to overcome critical optimization challenges.

%% file: sections/appendix.tex
\appendix
\section{Experiments Hyperparameters}
\label{appendix:hyperparameters}
\begin{table}[h]
\centering
\caption{Training Hyperparameters}
\label{tab:hyperparameters}
\begin{tabular}{lr}
\toprule
\textbf{Parameter} & \textbf{Value} \\
\midrule
\multicolumn{2}{c}{\textbf{SFT Training Parameters}} \\
\midrule
Per Device Train Batch Size & 4 \\
Hardware Configuration & 8$\times$8 H100 GPUs \\
Gradient Accumulation Steps & 8 \\
Learning Rate & $5 \times 10^{-6}$ \\
LR Schedule & Warmup-stable-decay \\
LR Warmup Steps & 200 \\
LR Min Value & $1 \times 10^{-6}$ \\
LR Decay Period & Final 10\% of steps \\
Number of Epochs & 100 \\
\midrule
\multicolumn{2}{c}{\textbf{RL Sampling Parameters}} \\
\midrule
RL Online Sampling Generation Length $L$ & 256 \\
Diffusion Steps & 128 \\
Block Length & 32 \\
Sampling Temperature & 1.2 \\
Generations Per Group $G$ & 8 \\

\midrule
\multicolumn{2}{c}{\textbf{RL Training Parameters}} \\
\midrule
Per Device Train Batch Size & 8 \\
Hardware Configuration & 8$\times$8 H100 GPUs \\
Gradient Accumulation Steps & 1 \\
Effective Batch Size & 512 \\
KL Beta $\beta$ & 0.01 \\
Policy Iterations $\mu$ & 4 \\
Learning Rate & $5 \times 10^{-7}$ \\
LR Schedule & Linear decay to 0 \\
LR Warmup Steps & 50 \\
LR Decay Period & 10 epochs \\
Training Steps & 1440 \\
Clip Ratio Epsilon $\varepsilon$ & 0.2 \\
\midrule
\multicolumn{2}{c}{\textbf{IGPO Specific Parameters}} \\
\midrule
Chunk Size $|c_j| \sim \mathcal{U}[s_{\text{min}}, s_{\text{max}}]$ & $\mathcal{U}[5, 10]$ \\
Inpainting Ratio $\eta_i \sim \mathcal{U}[\eta_{\text{low}}, \eta_{\text{high}}]$ & $\mathcal{U}[0.2, 0.6]$ \\
replacement fraction $\lambda$ & 0.5 \\
Entropy-based Gradient Filtering for Inpainted Tokens $\tau$ & 0.2 \\
\bottomrule
\end{tabular}
\end{table}
\newpage


\input{sections/igpo_inpaint_example}

\newpage
\section{Length-Aligned SFT: SFT trace revision length distribution comparison}
\label{appendix:sft_revision_token_length}
\begin{figure}[h]

    \centering
    \includegraphics[width=0.8\textwidth]{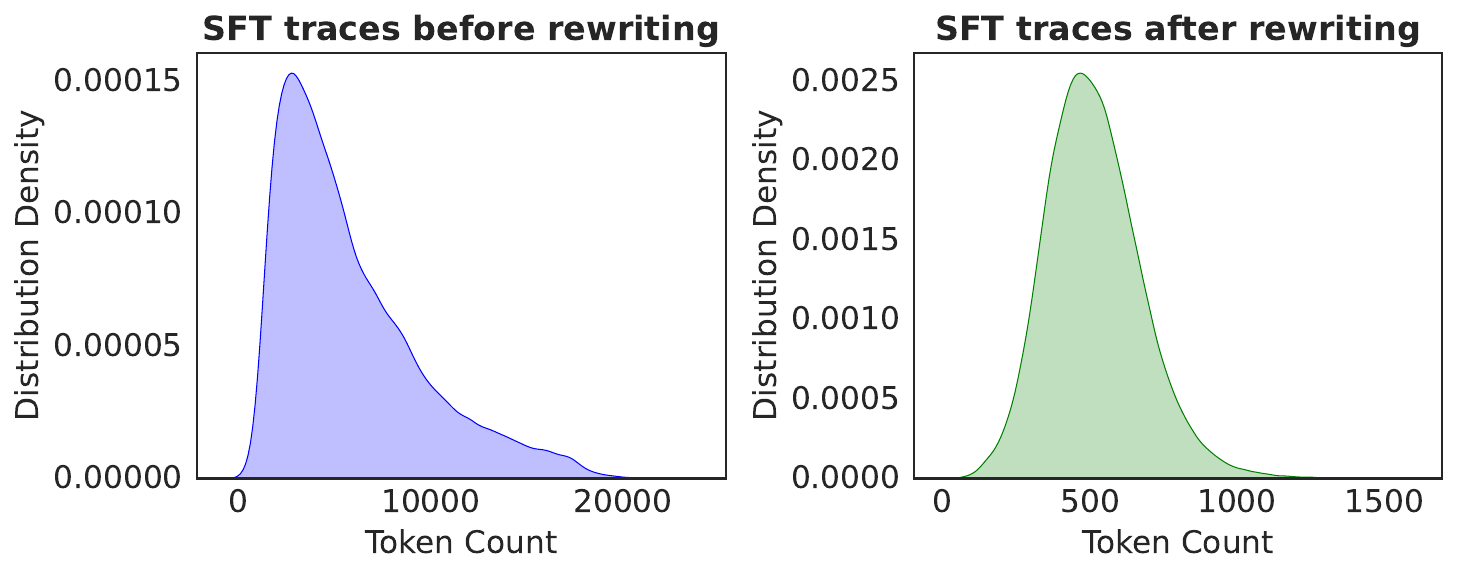}
   \caption{\textbf{Token Length Distribution of SFT Dataset Before and After Revision.} Comparison of token length distributions for the OpenR1-Math-220K dataset (94k math problems). After revision using LLaMA-4-Maverick, token lengths are constrained to below 1500 tokens, eliminating the extreme range of the original dataset where traces could exceed 20,000 tokens. This addresses the generation length mismatch across SFT training, RL sampling (256 tokens), and evaluation (512 tokens) phases.}
    \label{fig:token_length_sft}
\end{figure}

As illustrated in \cref{fig:token_length_sft}, the original OpenR1-Math-220K dataset exhibits substantial token length diversity, with reasoning traces extending beyond 10,000 tokens while LLaDA's maximum context length is only 4096 tokens. Naively applying SFT on this dataset would result in many truncated sequences, and even for samples within the 4096-token limit, significant distribution mismatch persists across training phases—we use 256 tokens for RL sampling and 512 tokens for evaluation. Our rewriting using LLaMA-4-Maverick successfully constrains all traces to under 1500 tokens, creating alignment between SFT training, RL sampling, and evaluation phases. Additionally, while reflective behavior has been found helpful for LLaDA in prior work~\citep{zhao2025d1}, the excessive repeated reflective patterns in the original dataset are unsuitable for its constrained generation space. The rewriting process eliminates this redundancy while preserving essential reasoning structure.
\newpage
\section{Prompt for SFT traces revision}
\label{appendix:revision_prompt}

\begin{tcolorbox}[float, breakable, title={Prompt for SFT traces revision sent to LLaMA 4 Maverick}]
Please rewrite the original solution to make it more concise and easier to understand without changing the details. Please put the explanation in the solution between \texttt{<reasoning>} and \texttt{</reasoning>} and put the final numerical answer between \texttt{<answer>} and \texttt{</answer>} in boxed format. Please shorten or rewrite the rewritten solution to a random length between 100 and 1000 words while keeping sufficient details of the reasoning steps. Please do not return anything other than the rewritten solution. 

\textbf{Example:} 
\begin{verbatim}
<reasoning>
xxx
</reasoning>

<answer>
\boxed{14}
</answer>
\end{verbatim}

\textbf{Original solution:} \{generations\}. 

\textbf{Your response:}
\end{tcolorbox}

\begin{tcolorbox}[float, breakable, title={\textbf{SFT trace before rewritten by LLaMA 4 Maverick}\\
\textbf{Problem:} Find all primes between 1 and 15. }]

$<$think$>$

Okay, so I need to find all the prime numbers between 1 and 15. Hmm, let's start by remembering what a prime number is. From what I recall, a prime number is a natural number greater than 1 that has no positive divisors other than 1 and itself. So, numbers like 2, 3, 5, etc., are primes because you can't divide them evenly by any other number besides 1 and the number itself. On the other hand, numbers like 4, 6, 8, and so on are not primes because they can be divided by other numbers. For example, 4 can be divided by 2, so it's not prime. 

Alright, so the task is to list all primes between 1 and 15. Let me list out all numbers from 1 to 15 first to check them one by one. The numbers are: 1, 2, 3, 4, 5, 6, 7, 8, 9, 10, 11, 12, 13, 14, 15.

Now, let's go through each number and see if it's prime. Starting with 1. \textcolor{blue}{Wait, I remember that by definition, prime numbers are greater than 1.} So 1 is not considered a prime. So we can skip 1.

Next up is 2. Let's check if 2 is prime. The divisors of 2 are 1 and 2. There are no other divisors, so yes, 2 is a prime number.

Moving on to 3. Similarly, the divisors are 1 and 3. No other numbers divide into 3, so 3 is prime.

Next is 4. Let's check divisors. 4 divided by 2 is 2, which is an integer, so 4 is not prime. It can be divided by 2, so it's composite.

Then 5. The divisors of 5 are 1 and 5. No other numbers divide into 5, so 5 is prime.

6. Checking divisors. 6 divided by 2 is 3, so it's divisible by 2 and 3. Not prime.

7. Divisors are 1 and 7. No other divisors, so 7 is prime.

8. Divided by 2 gives 4, so 8 is not prime.

9. Divided by 3 gives 3, so 9 is not prime (since 3 times 3 is 9).

10. Divided by 2 is 5, so 10 is not prime.

11. Let's see, 11. Divided by 2? 11 divided by 2 is 5.5, not an integer. Divided by 3? 11 divided by 3 is approximately 3.666... Not an integer. Divided by 4? 2.75, nope. 5? 2.2, nope. Since 11 is a prime number because none of these divisions result in integers, so 11 is prime.

12. Divided by 2 is 6, so composite.

13. Checking divisors. 13 divided by 2 is 6.5, 13 divided by 3 is about 4.333..., 4? 3.25, 5? 2.6, 6? 2.166... None of these are integers. So 13 is prime.

14. Divided by 2 is 7, so composite.

15. Divided by 3 is 5, so composite.

Let me recap the numbers between 1 and 15 and their primality:

1: Not prime

2: Prime 

3: Prime

4: Not

5: Prime

6: Not

7: Prime

8: Not

9: Not

10: Not

11: Prime

12: Not

13: Prime

14: Not

15: Not

So compiling the primes from this list: 2, 3, 5, 7, 11, 13.

\textcolor{blue}{Wait a second, let me verify each one again just to be sure I didn't make a mistake.}

Starting with 2: Definitely prime. It's the only even prime.

3: Prime, yes.

5: Prime, yes.

7: Prime. Correct.

11: Let's check again. 11 divided by 2 is not whole. 3? 3*3 is 9, 3*4 is 12. So no. 5? 5*2 is 10, 5*3 is 15. So 11 is prime.

13: Similarly, 13 divided by 2 is 6.5, 3 is 4.333, 5 is 2.6. So, prime.

Yes, those all check out. So primes between 1 and 15 are 2, 3, 5, 7, 11, and 13. 

 \textcolor{blue}{But hold on,} sometimes people might make a mistake with 9. \textcolor{blue}{Wait, 9 is 3 times 3, so that's composite, which I had down.} 15 is 3 times 5, composite. So all the non-primes are accounted for. 

Is there any number I might have missed? Let's see. Let's list all numbers again:

2, 3, 5, 7, 11, 13. Are there any others?

Well, 1 is excluded. 4,6,8,9,10,12,14,15 are composite. So no, those six numbers are all primes between 1 and 15.

\textcolor{blue}{Wait, but the question says "between 1 and 15." Does that include 15?} Hmm, the wording is "between 1 and 15," which could be a bit ambiguous. But usually, when someone says "between 1 and 15," they mean including both endpoints. However, since 15 is composite, it's excluded either way. So primes are 2,3,5,7,11,13.

Yes, that seems correct. I don't think I made any mistakes here. Let me cross-verify with another method, like the Sieve of Eratosthenes. The Sieve is a method where you list all numbers and cross out multiples.

Starting with 2, cross out multiples: 4,6,8,10,12,14.

Then 3, cross out multiples: 6,9,12,15.

Then 5, cross out 10,15.

7, cross out 14.

11 and 13 are primes, and their multiples are beyond 15. So remaining numbers are 2,3,5,7,11,13. Which matches our previous list.

Therefore, the primes between 1 and 15 are 2,3,5,7,11,13.

**Final Answer**

The primes between 1 and 15 are \boxed{2}, \boxed{3}, \boxed{5}, \boxed{7}, \boxed{11}, and \boxed{13}.

$<$/think$>$

To find all the prime numbers between 1 and 15, we start by recalling that a prime number is a natural number greater than 1 that has no positive divisors other than 1 and itself. We then check each number from 1 to 15:

- **1**: Not a prime number (by definition).
 
- **2**: Prime (divisors: 1, 2).

- **3**: Prime (divisors: 1, 3).

- **4**: Not prime (divisible by 2).

- **5**: Prime (divisors: 1, 5).

- **6**: Not prime (divisible by 2, 3).

- **7**: Prime (divisors: 1, 7).

- **8**: Not prime (divisible by 2).

- **9**: Not prime (divisible by 3).

- **10**: Not prime (divisible by 2, 5).

- **11**: Prime (divisors: 1, 11).

- **12**: Not prime (divisible by 2, 3, 4, 6).

- **13**: Prime (divisors: 1, 13).

- **14**: Not prime (divisible by 2, 7).

- **15**: Not prime (divisible by 3, 5).

Using the Sieve of Eratosthenes method to cross out non-prime numbers confirms the primes listed above. Therefore, the prime numbers between 1 and 15 are:

\[
\boxed{2}, \boxed{3}, \boxed{5}, \boxed{7}, \boxed{11}, \boxed{13}
\]

\end{tcolorbox}

\begin{tcolorbox}[float, breakable, title={\textbf{SFT trace after rewritten by LLaMA 4 Maverick}\\
\textbf{Problem:} Find all primes between 1 and 15. }]

 $<$reasoning$>$ 
 
To find all the prime numbers between 1 and 15, we recall that a prime number is a natural number greater than 1 that has no positive divisors other than 1 and itself.

We then check each number from 1 to 15 to determine if it is prime.

- **1**: Not a prime number (by definition).

- **2**: Prime (divisors: 1, 2). 

- **3**: Prime (divisors: 1, 3). 

- **4**: Not prime (divisible by 2). 

- **5**: Prime (divisors: 1, 5). 

- **6**: Not prime (divisible by 2, 3). 

- **7**: Prime (divisors: 1, 7). 

- **8**: Not prime (divisible by 2). 

- **9**: Not prime (divisible by 3). 

- **10**: Not prime (divisible by 2, 5). 

- **11**: Prime (divisors: 1, 11). 

- **12**: Not prime (divisible by 2, 3, 4, 6).

- **13**: Prime (divisors: 1, 13). 

- **14**: Not prime (divisible by 2, 7). 

- **15**: Not prime (divisible by 3, 5).

The Sieve of Eratosthenes method confirms the primes listed above.  

$<$/reasoning$>$

$<$answer$>$ 

\boxed{2, 3, 5, 7, 11, 13} 

$<$/answer$>$

\end{tcolorbox}

%% file: sections/igpo_inpaint_example.tex
\section{Inpainting Generation Qualitative Examples}
\label{appendix:inpainted_examples}

\begin{tcolorbox}[
    float, 
    breakable, 
    enhanced,
    colframe=gray!40,
    colback=white,
    coltitle=black,
    colbacktitle=gray!20,
    title={
           \small{\textcolor{blue!70!black}{\textbf{Blue text:}} Question Prompt} \quad 
           \textcolor{ForestGreen}{\textbf{Green text:}} Injected Hints \quad 
           \textcolor{black}{\textbf{Black text:}} Inpainted Generation}
]
\footnotesize
\textbf{Problem:} Circle $C$ has radius 6 cm. How many square centimeters are in the area of the largest possible inscribed triangle having one side as a diameter of circle $C$?
\vspace{0.3cm}
\hrule
\vspace{0.3cm}
\textbf{Inpainting Input At Denosing Step 0:}
\begin{tcolorbox}[
    enhanced,
    colframe=gray!60,
    colback=gray!10,
    boxrule=0.5pt,
    arc=2pt
]
{\footnotesize
\begin{alltt}
\textcolor{blue!70!black}{<|start_header_id|>user<|end_header_id|>
Respond in the following format:
<reasoning>
...
</reasoning>
<answer>
\textbackslash boxed\{<Your answer>\}
</answer>
Circle \$C\$ has radius 6 cm. How many square centimeters are in the area 
of the largest possible inscribed triangle having one side as a diameter of 
circle \$C\$?<|eot_id|><|start_header_id|>assistant<|end_header_id|>}

\textcolor{gray!60}{<|mdm_mask|><|mdm_mask|>...<|mdm_mask|>}
\textcolor{ForestGreen}{\textbf{Since the diameter is twice the radius, the base of the triangle will have 
length \$2 \textbackslash times 6 = 12\$ cm.
To}} \textcolor{gray!60}{<|mdm_mask|>...<|mdm_mask|>} \textcolor{ForestGreen}{\textbf{, we want the height to be as large}} \textcolor{gray!60}{<|mdm_mask|>...<|mdm_mask|>} 
\textcolor{ForestGreen}{\textbf{of the triangle is the distance from the midpoint of the base to the 
circumference of the circle.
This is equal to the radius of the circle,}} \textcolor{gray!60}{<|mdm_mask|>...<|mdm_mask|>} 
\textcolor{gray!60}{<|mdm_mask|>...<|mdm_mask|>} \textcolor{gray!60}{<|mdm_mask|>...<|mdm_mask|>} 
\textcolor{ForestGreen}{\textbf{12 \textbackslash times}} \textcolor{gray!60}{<|mdm_mask|>...<|mdm_mask|>}
\end{alltt}
}
\end{tcolorbox}
\vspace{0.2cm}
\textbf{Inpainted Generation After Denoising:}
\begin{tcolorbox}[
    enhanced,
    colframe=green!60,
    colback=green!5,
    boxrule=0.5pt,
    arc=2pt
]
{\footnotesize
\begin{alltt}
<reasoning>
We know to find the largest possible inscribed triangle in circle \$C\$, the base 
of the triangle should be the diameter of the circle.
\textcolor{ForestGreen}{\textbf{Since the diameter is twice the radius, the base of the triangle will have 
length \$2 \textbackslash times 6 = 12\$ cm.
To}} find the largest possible area of the triangle\textcolor{ForestGreen}{\textbf{, we want the height to be as large}} 
as possible.
The largest possible height \textcolor{ForestGreen}{\textbf{of the triangle is the distance from the midpoint of the base to the 
circumference of the circle.
This is equal to the radius of the circle,}} which is 6 cm.
Therefore, the area of the largest possible inscribed triangle is given by the 
formula for the area of a triangle, which is \$\$\textbackslash frac\{1\}\{2\} \textbackslash times \textbackslash text{base} 
\textbackslash times \textbackslash text{height} = \textbackslash frac\{1\}\{2\} \textbackslash times \textcolor{ForestGreen}{\textbf{12 \textbackslash times}} 6 = 36\$\$ square centimeters.
</reasoning>
<answer>
\textbackslash boxed\{36\}
</answer>
\end{alltt}
}
\end{tcolorbox}
\end{tcolorbox}